\documentclass[a4paper,fleqn]{cas-sc}
\usepackage{caption,setspace}
\usepackage[authoryear,longnamesfirst]{natbib}
\usepackage{graphicx}

\begin{document}
\let\WriteBookmarks\relax
\def\floatpagepagefraction{1}
\def\textpagefraction{.001}

\shorttitle{Inappropriate content detection in Urdu Language}

\shortauthors{E Shoukat et~al.}

\title [mode = title]{Attention based Bidirectional GRU hybrid model for inappropriate content detection in Urdu language.}                      
\tnotemark[1]
\tnotetext[1]{This work is not sponsored/supported by any organization.}


%
\author[1]{E Shoukat}[type=editor,
                        ]



\ead{eshoukat.msit18seecs@seecs.edu.pk}


\credit{Methodology, Data curation, Writing - Original draft preparation, Visualization}

\affiliation[1]{organization={School of Electrical Engineering and Computer Science (SEECS), National University of Sciences and Technology (NUST)},
    addressline={H-12}, 
    city={Islamabad},
    country={Pakistan}}

\author[1]{R Irfan}
                    [
                        orcid=0000-0002-2719-9852
                        ]
\cormark[1]

\fnmark[1]

\ead{rabia.irfan@seecs.edu.pk}

\credit{Formal analysis, Supervision, Writing – review \& editing, Project administration, Validation}
\author[1]{I Basharat}
\cormark[1]
\fnmark[1]
\ead{ibasharat.phdcs20seecs@seecs.edu.pk}

\credit{Review \& editing, Validation}
\author[1]{MA Tahir}
                    [
                        orcid=0000-0002-2335-2776
                        ]
\ead{ali.tahir@seecs.edu.pk}


\credit{Conceptualization, Formal analysis, Supervision, Writing – review \& editing, Project administration, Validation}

\author[2]{S Shaukat}
                    [
                        ]
\fnmark[2]
\ead{sameen.shaukat@yahoo.com}

\affiliation[2]{organization={University of Illinois Urbana-Champaign}, 
    country={United States}}
\credit{Formal analysis, Visualization, Validation, Project administration }



\begin{abstract}
With the increased use of the internet and social networks for online discussions, the spread of toxic and inappropriate content on social networking sites has also increased. Several studies have been conducted in different languages. However, there is less work done for South Asian languages for inappropriate content identification using deep learning techniques. In Urdu language, the spellings are not unique, and people write different common spellings for the same word, while mixing it other languages, like English in the text makes it more challenging, and limited research work is available to process such language with the finest algorithms. The use of attention layer with a deep learning model can help handling the long-term dependencies and increase its efficiency . To explore the effects of the attention layer, this study proposes attention-based Bidirectional GRU hybrid model for identifying inappropriate content in Urdu Unicode text language. Four different baseline deep learning models; LSTM, Bi-LSTM, GRU, and TCN, are used to compare the performance of the proposed model. The results of these models were compared based on evaluation metrics, dataset size, and impact of the word embedding layer. The pre-trained Urdu word2Vec embeddings were utilized for our case. Our proposed model BiGRU-A outperformed all other baseline models by yielding 84\% accuracy without using pre-trained word2Vec layer. From our experiments, we have established that the attention layer improves the model's efficiency, and pre-trained word2Vec embedding does not work well with an inappropriate content dataset.
\end{abstract}




\maketitle

\section{Introduction}
Social media networks like Twitter, Facebook, and YouTube allow a wide diversity of people from all around the world. Users find it more convenient to write and express their thoughts about products, movies, or articles in their local language than in English \cite{akhter2020automatic}. At the same time, billions of people using social media platforms are prone to cyber-crimes such as bullying, threats, and scams. The Government of Pakistan (GoP) has taken precautionary steps to avoid uncontrollable consequences after a series of events in recent years \cite{rizwan2020hate}. These included slandering of political parties, and their leaders, hurting the sentiments of religious minorities, and targeted harassment of women sharing their point of view. Such actions give away the notions of nation's struggles with online hate speech dialogues and necessitates the immediate need for automated filtering system.\\
Social media sites need to be able to detect, categorize, and clean the inappropriate content before it is posted online. Inappropriate content in this study is used broadly to refer to any type of the content that could be hurtful. Abusive language and profanities are major part of the inappropriate or violating content. Urdu is a resource-scarce language with a complex morphological structure, unique characters, and low linguistic resources \cite{mandl2019overview}. Several native Urdu speakers communicate their sentiments, views, and thoughts on social media using the Urdu script. Hence, automatic filtering and identification of inappropriate content in the Urdu language are equally important to help non-Urdu speakers understand the sentiments and to improve the quality of conversation for native Urdu speakers.\\
Many studies been conducted using Machine learning (ML) and Deep learning (DL) algorithms to detect inappropriate content in different languages \cite{chakraborty2019threat},\cite{Sigurbergsson2020OffensiveLA},\cite{subramani2019deep}. From basic ML models to simple recurrent neural networks to the most advanced models with transformers are the approaches followed by researchers. In this domain, much work been done in English and many other languages. But current literature lacks inappropriate content detection in Urdu script using DL techniques. In recent years many researchers have publicly shared their annotated Urdu corpus for future research workers. However, finding domain-specific data is still a challenging task. Most of the annotated datasets available are either in Roman Urdu or of small size to apply advanced deep neural networks and analyze their results. Furthermore, only classical ML models are explored extensively on the available datasets.\\
The proposed strategy aims to address the gaps discussed till yet. The following are the key contributions of this research:
\begin{itemize}
  \item A large inappropriate content corpus, developed by combining three publicly available annotated datasets. The first two were in Urdu Script, but the third was converted from Roman Urdu to Urdu script.
  \item A Bidirectional Gated Recurrent Unit with attention layer (BiGRU-A) hybrid model for identification of inappropriate content is proposed.
  \item Result comparison of state-of-art DL models, i.e. Long short-term Memory (LSTM), Bidirectional Long short-term Memory (Bi-LSTM), Gated Recurrent Unit (GRU), Temporal Convolutional Network (TCN).
  \item Comparative analysis of baseline models and proposed model with and without using pre-trained word embedding model word2vec for the Urdu language.
  \item The impact of dataset size on the performance of DL models is studied.
\end{itemize}
The paper is comprised of following sections; introduction followed by the study of existing literature, research gap, problem statement, methodology, experiments and discussion on their results, conclusions, future work and references.
\section{Literature Review}
This section provides in-depth information on studies related to the identification of inappropriate language. It explores conventional and neural network approaches. Finally, we will provide a detailed comparative analysis of the work on offensive content detection in the Urdu language. Previous studies have explored classical ML models and a few Neural Networks(NN) for the problem of offensive language detection in Urdu script. On large data sets, ML models struggle to perform well and usually learn just directed text features. When compared to standard ML models, the efficiency of DL models has significantly improved as the size of dataset increases \cite{mathew2020deep}. The NLP community has been concentrating on internet platforms including Twitter,YouTube, Instagram, Facebook, and online blogs to identify harmful language \cite{Balakrishnan2019CyberbullyingDO}, \cite{Alakrot2018TowardsAD}, \cite{rani2019kmi}, \cite{Lee2018AnAT}.
\subsection{Machine Learning approaches}
In \cite{chakraborty2019threat}, Support Vector Machine(SVM) with the linear kernel is proposed to achieve the best results in detecting inappropriate language in Bengali dataset. In their suggested approach, both Unicode the Bengali characters, as well as Unicode emoticons, were taken into consideration as acceptable input in our suggested approach. Authors in \cite{hajibabaee2022offensive} employed ML classifiers SVM, Logistic Regression (LR), Decision Trees (DT) and Random forest (RF), etc. along with modular cleaning of Twitter dataset, and built a tokenizer. Threatening comments in Twitter tweets were categorized using NB in \cite{lee2018comparative}. Multi-class classification in Roman Urdu content extracted from YouTube comments is presented by \cite{9318069}. For feature selection, n-gram and TF-IDF techniques are applied first. For normalization, L1 and L2 approaches are applied. They also used SMOTE to balance classes since some labels had fewer instances. For comparative analysis , LR, Naive Bayes (NB), SVM, and Stochastic Gradient Descent (SGD) classifier are utilized. Their findings demonstrated that SVM combined with n-gram feature selection, L2 normalization and TF-IDF feature values outperforms all other models on their dataset. The hyperparameters of ML models were also tuned using 10-fold cross-validation. Additionally, they created a web interface for YouTube Monitor, firstly, to scrape user comments from a keyword or given URL and then classify them into respective hate content categories.
\subsection{Deep Learning approaches}
A hybrid deep learning model by combining Convolutional Neural Network (CNN) and Bi-LSTM is proposed in this study \cite{yenala2018deep} for automatically identifying inappropriate language. They were particularly interested in finding a solution for the following two application scenarios:(a) search engine query completion recommendations,and (b) chats of user in messenger.In \cite{zimmerman2018improving} authors used an ensemble method to advance the application of DL models for hate speech content identification. Most of the posts in the datasets used for hate content identification are extracted from social media sites like Twitter. When compared to other categories, the proportion of hate speech incidents in the real world was relatively low. This distribution between hate speech and different categories can be concluded from the majority of the statistics gathered from social media.  The lack of hate speech instances in datasets is recognized to be a difficulty for its detection tasks. Various NLP tasks can benefit from transfer learning approaches, including universal language model fine-tuning (ULMFiT), generative pre-trained transformer (GPT), embedding from language models (ELMO), and bidirectional encoder representations from transformers (BERT) \cite{yadav2021comparative}.\\
The issue of identifying the inappropriate language in Dravidian languages, i.e. Malayalam, Tamil, and Kannada taken from Youtube comments are viewed in this study \cite{andrew2021judithjeyafreedaandrew} as a multiclass classification problem. The study then presents the accuracy estimates for ML models on the training data. They have established an inflection point with significant advancements, particularly in activities where data is scarce. To increase the effectiveness of hate speech identification, the authors tested various fine-tuning techniques. The authors have concluded that the CNN-based model and BERT transfer learning models can perform better than other approaches that been looked at.
\subsection{Approaches on Urdu language dataset}
The recently released study by \cite{rizwan2020hate} is possibly the most important work in detecting offensive language in the Roman Urdu dataset. The researchers have presented their findings for both coarse-grained and fine-grained classification tasks using a variety of widely used baseline ML and DL models. They demonstrated how their unique BERT and CNN-ngram hybrid model might be used for transfer learning and attain an outclass F1 score on a coarse-grained classification problem. \cite{amjad2021threatening} provides details on how to recognize threatening language and identify targets in Twitter posts written in Urdu. The authors of this research offered a dataset that consists of 3,564 Twitter messages that been manually classified as either harmful or non-harmful by human specialists. The target further categorises the threatening tweets into one of two categories: threats against an individual or threats against a group. Numerous experiments using various ML and DL techniques revealed that the best threatening content detection accuracy was achieved by a Multilayer Perceptron (MLP) classifier combined with the word n-gram model. Whereas, SVM, along with fastText word embedding, outperforms for target identification tasks.\\
In this study, \cite{ali2022hate}, they formulated an annotated hate speech lexicon for the Urdu language of 10,526 tweets. They also employed various ML methods for the detection of hate content as baseline experiments. Additionally, applied  transfer learning approaches to make use of multilingual BERT, and FastText Urdu word embeddings for their assignment. They tested four alternative BERT versions and achieved encouraging results for multi-class classification problems.
\subsection{Comparative Analysis}
As already established, when it comes to the identification of hate speech content, a lot of the study is focused on the English language. All feature extraction methods, pattern discovery procedures, and models are consequently built for English. In addition, there is a dearth of structured data that may be used for research in poor resource languages. For English, there are some quite large annotated publicly available datasets, but the same cannot be stated for resource-scarce languages like Urdu for this problem space. Additionally, to the best of our knowledge, there isn't any extensive published research on the identification of inappropriate content in the Urdu language using deep learning algorithms at this time. Due to this, it is currently difficult to determine which method would work best for a given dataset or how the training data should be modified to derive the best predictions.\\
We summarize the existing literature related to inappropriate language detection in Table 1. The first column lists all the papers from which the literature review is conducted. The next column describes the platform from where the dataset is collected. From the literature it was observed that most of the hate speech content data sources are social media sites. As large community of people having backgrounds from all around the world has easy access to social media sites. Hence, these sites are the best source for collecting data related to inappropriate content detection domain.\\
The "Features" column lists most commonly used feature extraction techniques. TF-IDF is a reasonably straightforward but intuitive method of weighing words, making it a perfect starting point for several tasks. For the deep learning model, use of word embeddings is recommended. Word2Vec is the most commonly used word embedding. It is also available for many languages, including Urdu. The following two columns describe that most of the studies are conducted in English, and very few in Urdu specifically, Urdu Nastaliq script. And those research that analyses the Urdu script either have extremely tiny dataset sizes to investigate deep learning methods or have a very low number of words in each label to extract the true features effectively. Lastly, different ML and DL models are listed in the last column.
\begin{table}[width=.9\linewidth,cols=7]
\caption{Comparative Analysis of Existing Work.}\label{tbl1}
\begin{tabular*}{\tblwidth}{p{1.5cm}p{1.5cm}p{2cm}p{1.5cm}p{1.5cm}p{3cm}p{1cm}}
\textbf{Paper} & \textbf{Platform} & \textbf{Features} & \textbf{Language} & \textbf{Data size} & \textbf{Technique} & \textbf{Evaluation} (\%)\\
\midrule
\cite{chakraborty2019threat} & Facebook & TF-IDF, n-gram & Bengali & Small & MNB, SVM, CNN-LSTM & 78\\
\cite{hajibabaee2022offensive} & Twitter & TF-IDF, W2V, FastText & English & Large & MLP, SVM, RF, LR, GB, DT, AdaBoost, NB & 95(F1)\\
\cite{lee2018comparative} & Twitter & TF-IDF, n-grams & English & Large & NB, LR, SVM, RF, GBT, CNN, RNN & 80.5(F1)\\
\cite{rani2019kmi} & Twitter & n-gram & English & Moderate & linear SVM, DT & 79.76 \\
\cite{akhter2020automatic} & Twitter & n-gram & Urdu, Roman Urdu & Moderate & 17 ML algos & 95.8(F1)\\
\cite{9318069} & YouTube comments & TF-IDF, n-gram, L1, L2  & Roman Urdu & Moderate & LR, SVM, SGD, NB & 77.45\\
\cite{yenala2018deep} & search engine, messenger & -  & English & Large & LSTM, C-BiLSTM, BLSTM & 92\\
\cite{subramani2019deep} & Facebook & W2V, GloVe  & English & Large & CNN, RNN, GRU, LSTM, BLSTM & 87(F1)\\
\cite{zimmerman2018improving} & Twitter & Keras EL & English  & Moderate & Ensemble CNN & 5\% increased (F1)\\
\cite{andrew2021judithjeyafreedaandrew} & YouTube comments & TF-IDF & Dravidian  & Large & NB, SVM, KNN, DT, LR, RF & 93(F1)\\
\cite{rizwan2020hate} & Twitter & FastText, BERT & Roman Urdu & Moderate & hybrid DL models, CNN-gram & 82\\
\cite{amjad2021threatening} & Twitter & n-gram, FastText  & Urdu & Small & LR, RF, AdaBoost, MLP, SVM, 1D-CNN, LSTM & 75.31\\
\cite{ali2022hate} & Twitter & TF-IDF,CV, w2v,FastText, BERT  & Urdu & Moderate & ML, CNN, BiGRU & 69(F1)\\
\bottomrule
\end{tabular*}
\end{table}
\subsection{Research Gap}
The following research gaps were identified after an extensive literature review and have been addressed in our study.
\begin{itemize}
    \item The quality of the conversation by automatic filtering of inappropriate content in Urdu script using deep learning algorithms.
    \item Identification of inappropriate content in the Urdu script.
    \item Exploring advanced deep learning model with attention layer.
    \item Comparison with baseline DL models
    \item Impact of word embeddings on inappropriate content dataset.
\end{itemize}
We proposed an attention-based bidirectional GRU hybrid model for identifying inappropriate content in Urdu script. Our research seeks to contribute by developing and making available an extensive dataset in Urdu, comparing the effectiveness of embedding features, studying the impact of dataset size, and evaluating DL models to assess their effectiveness. 
\section{Dataset}
This section illustrates the process of data collection and statistics of the data. It is constituted from the literature review that Urdu is a resource-scarce language. To collect domain specific data for such languages is the foremost important and difficult task.
\subsection{Dataset collection}
Three publicly available datasets in Urdu native from the online Internet source are obtained for this problem space.
\begin{itemize}
    \item The first two datasets\footnote{\label{note1}\url{https://github.com/MaazAmjad/Threatening_Dataset}} \footnote{\label{note2}\url{https://github.com/pervezbcs/Urdu-Abusive-Dataset}} are obtained using twitter Application Programming Interface (API) with tweets containing violent or abusive content labeled ``1'' and neutral content labelled as ``0''.
    \item Third dataset\footnote{\label{note3}\url{https://github.com/haroonshakeel/roman_urdu_hate_speech}} obtained was in Roman Urdu language. It has same labels identifying inappropriate content as ``1'' and neutral as ``0''. This dataset is first converted from Roman Urdu to Urdu script using online website called ijunoon\footnote{http://www.ijunoon.com/}.
\end{itemize}
Finally our dataset is formed with the combination of these three datasets. To study the impact of size of dataset, the dataset is partitioned in variable sized groups. It is combined to form two groups i.e One is UrduInAsmall and other is UrduInAlarge. UrduInAsmall is formed by combining first two datasets\footref{note1} \footref{note2} and UrduInAlarge is the combination of all three datasets.\footref{note1} \footref{note2} \footref{note3} \\
Both data sets are have two categories i-e Inappropriate they are labeled as `1'and Appropriate that are labeled as `0'. Table 2 shows the two categories and its labels. A sample of dataset presented in figure 1 where the types of text in two classes can be understood easily.
\begin{table}[width=.9\linewidth,cols=3,pos=h]
\caption{Dataset categories.}\label{tbl2}
\begin{tabular*}{\tblwidth}{@{} LLL@{} }
\toprule
\textbf{Class} & \textbf{Label} & \textbf{Label Name}\\
\midrule
Class A & 1 & Inappropriate \\
Class B & 0 & Appropriate \\
\bottomrule
\end{tabular*}
\end{table}
\begin{figure}
	\centering
		\includegraphics[scale=.95]{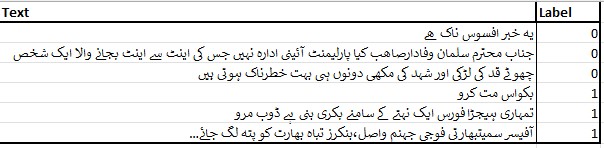}
	\caption{Sample of Dataset.}
	\label{FIG:1}
\end{figure}
\subsection{Dataset Statistics}
In UrduInAsmall, there are total 5734 entries of tweets from which 2890 instances are categorized as Inappropriate class and other 2844 are labelled as Appropriate class. In UrduInAlarge, there are 14946 text instances in our dataset that are divided into two classes i-e Inappropriate and Appropriate. 
From 14946 instances of total classes there are 7181 tweets in Inappropriate class, they are labeled as `1' and 7765 items in Appropriate class that are labeled as `0'. Table 3 shows the statistics of two data sets.
\begin{table}[width=.9\linewidth,cols=3,pos=h]
\caption{Statistics of dataset.}\label{tbl3}
\begin{tabular*}{\tblwidth}{@{} LLL@{} }
\toprule
\textbf{Characteristics} & \textbf{UrduInAsmall} & \textbf{UrduInAlarge}\\
\midrule
Total lines & 5734 & 14946 \\
Inappropriate & 2890 & 7181\\
Appropriate & 2844 & 7765\\
Maximum words in sentence & 198 & 240\\
\bottomrule
\end{tabular*}
\end{table}
The comparison of two different-sized data sets aims to investigate how these datasets affect the performance of DL models.
\section{Methodology}
In this section we briefly discuss the methodology of our proposed model for identification of inappropriate content detection. DL has yet to be fully investigated for detection of inappropriate content in Urdu script. By using a hybrid DL strategy, the use of our suggested architecture and comparison with basic DL models tends to close the gap that has been identified in the studied literature.\\
The proposed model integrates elements of well-known neural network models, specifically the attention based Bi-directional GRU. Figure 2 is an illustration of the suggested Bi-GRU model with the attention mechanism. Both the Bidirectional GRU and the attention layer are well-known for their applications in the text classification, which is why this hybrid model was merged to evaluate the adaptability of the former with the latter. The Bi-GRU and the attention layer serve distinct functions during the text classification process. The bidirectional gated recurrent unit was implemented to manage the unique aspect of polarity in text classification data and obtain independent context semantic information in the forward and backward passes. The attention mechanism gives weights to the features based on how much they contribute to classification.
\begin{figure}
	\centering
		\includegraphics[scale=.55]{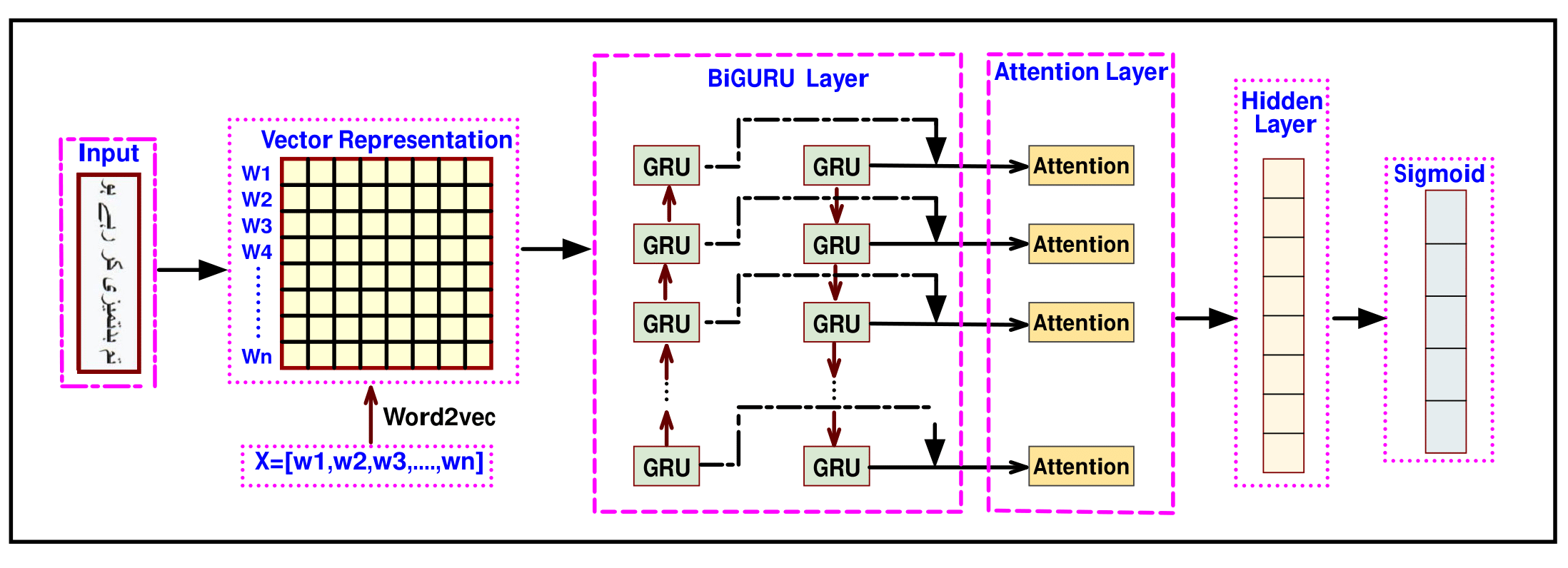}
	\caption{Attention-based Bidirectional Gated Recurrent Unit BiGRU - A Proposed model}
	\label{FIG:2}
\end{figure}
\subsection{Bidirectional GRU}
The suggested bidirectional RNN is utilized to handle the problem where the last output is not only associated to the preceding state, but also with the subsequent state. A Bi-RNN may learn both the forward and backward properties of the data. A forwarding and backward network combination like this will suit data better than a unidirectional RNN. Text classification frequently makes use of RNN, and while dealing with lengthy sequences, the standard RNN is exposed to the issues of vanishing gradient and explosion. The bidirectional GRU is a unique variation of the bidirectional RNN that divides the regular GRU into two directions: a forward direction associated with historical data and a reverse direction associated with future data. This allows the simultaneous use of the input observational data and future data. The unidirectional GRU's classification performance can be significantly enhanced by this configuration. Compared to RNN, it has significant advantages in handling long sequence texts and solves the gradient disappearance and explosion problems by adding update gate and reset gate to neurons. It also extracts text context information more successfully. This study represents a deep learning text classification technique based on a hybrid BiGRU attention model. The following list represents the main contents:
\begin{itemize}
\item[1.] Word embedding methodology to train the word vector, and the text data is encoded as a low-space dense matrix.
\item[2.] To extract text context characteristics, BiGRU is implemented.
\item[3.] The attention layer receives the output of BiGRU as an input to compute the attention score.
\end{itemize}
\subsection{Attention}
An attention mechanism is a part of a neural network and the essence of the attention process is the weight distribution of tokens. The more significant the words with higher weights are in the entire text, the more vital will be their role in the entire classification task. At each decoder stage, it judges what source elements are more important. The encoder in this arrangement does not have to vectorize the entire sentence; instead, it gives representations for each source token, such as the entire set of RNN states rather than just the most recent one. The basic idea is that a network can determine which input elements are more important at each level. Everything in this scenario is differentiable, allowing for a model based on attention to be trained from start to finish. You don't need to explicitly train the model to select the terms you want; it will figure out how to choose crucial information on its own and it is added individually to the attention layer. At each decoder step h\textsubscript{t}, attention receives input from all encoder states (s\textsubscript{1}, s\textsubscript{2}, s\textsubscript{3},..., s\textsubscript{t}), among others, and computes the attention scores. For the decoder state, attention determines the importance of each encoder state. In essence, it executes an attention function that receives input from a single encoder state and a single decoder state and produces a scalar value. The most often used techniques for determining attention scores are:
\begin{itemize}
\item[1.] The simplest way is dot-product.
\item[2.] To more efficient is to encode each source word, the encoder makes use of two RNNs that read input in the opposite directions; forward and backward.
\item[3.] The attention score is calculated using a bi-linear function and a uni-directional encoder.
\end{itemize}
Additionally, deep bidirectional GRU with attention layer offers more potent expressiveness and learning capabilities. The attention layer is a suggested technique for simplifying the modeling of long-term dependency. A more direct relationship between the model state at various periods in time is made possible by adding this layer\cite{raffel2015feed}.\\
In addition to our suggested model, GRU, LSTM, BiLSTM, and TCN have all been used on a curated dataset. The suggested method provides better classification accuracy for differently sized data sets.
\subsection{Word Embedding Vectors}
Neural network-based models known as "word embeddings" combine the information as dense and dispersed vectors. They hold the key to improving NLP results for resource-scarce languages. In addition to enhancing the performance of deep learning algorithms in numerous NLP applications, the use of word embedding has inspired critically needed research on these languages.\\
Words are represented in word embedding as vectors that encode their semantic characteristics. Word embeddings are frequently used as input characteristics in deep learning models for NLP \cite{collobert2011natural}. The process of language modelling and feature learning involves the transformation of tokens into vectors of progressive real values.\\
Matrix factorization and the usage of neural networks are two popular techniques for learning word embeddings \cite{mikolov2013efficient}. Word2Vec is a widely used word embedding system that efficiently trains word embeddings from text using a neural network prediction model. There are two models in it: the Skip-gram (SG) model and the Continuous Bag of Words (CBoW) model. In contrast to the SG model, which uses the target words to predict the context words, the CBoW model uses context words to predict the target word. The CBoW approach, which works well with tiny datasets, treats the full text context as a single observation. The SG model, however, treats each context and the target word pair as a separate observation and performs best with big datasets. For the Urdu language, pre-trained word embeddings can be created using the Word2Vec skip gram model \cite{kumhar2021word} which are used in this study. The context word vector problem can be solved by skip-garm model during training by calculating the conditional probability of the middle word vector. It is calculated as follows:
Let us assume that the total number of textual words fed into the model are \emph{k}
\begin{equation}
P (w_i | w_t ) = \frac {P(w_t * w_i)}{P(w_t)}
\end{equation}
where $w_{i}$ = text represented by vector $[w_1,w_2,...,w_k]$ and \emph{i = t - 1, t - 2, t + 1, t + 2.}\\
When it is passed through the word embedding layer, the new text representation becomes:
    \[x = [x_1, x_2, . . . , x_k ] \]
    where $x_{i} \in \mathbb{R}^{d}$ and \emph{d is the dimension of the word vector}
\section{Experimental setup}
An insight into data set pre-processing, configuration of hyper-parameters and evaluation metrics is given in this section. For our experiments we briefly explored the methods used while implementing the suggested model, and presented detailed comparison between the results obtained from proposed model, and other baseline models on two datasets. We have also addressed the identified research gaps and analyzed the outcome with the evaluation metrics used. We observed the performance of our model in both datasets and studied the impact of word2Vec word embedding.
\subsection{Data Prepossessing}
The first core task in every NLP problem is the pre-processing of  the dataset. 
It helps to organize the information by applying basic operations on it before it is ready to feed into a neural network. Other operations of the process include the removal of white spaces and unnecessary words, converting words into their root forms, elimination of redundant words, and tokenizing of the translated sentences and developing a lexicon for source languages. It transforms the raw dataset into a organized and meaningful dataset for further processing.\\
To get accurate results, a well organized, thoroughly cleaned, and normalized data is preferred. Pre-processing keeps data clean and free of noise and redundant information. Researchers use this method frequently to obtain cleaned data for improved model interpretation, thus we  have performed text pre-processing for standardisation of our datasets.\\
Though the pre-processing of the Urdu language is a laborious process by itself, however, an Urdu pre-processing library named Urduhack\footnote{https://pypi.org/project/urduhack/} has made it quite simple.
\subsection{Hyper-parameter Tunning}
While training the DL models choosing the best number of neurons in the dense layer, number of dense layer and optimization of parameters is tedious task for implementing a numbers of models to solve multiple tasks. The primary drawback is the potential for each model's training to proceed slowly. The models must be trained once for each potential set of parameters in order to determine the best parameters. Since it is difficult to test every possible combination, we move on to a few sets that ought to work well. To achieve best accuracy performance of all models to perfectly compare with the proposed models different combination of parameters, layers and neurons are utilized to finally conclude this study.\\
For our proposed architecture, after repeated experiments we concluded to use the parameters presented in Table 4. Activation function 'sigmoid' is used for for binary classification as the output of this function is always either '0' or '1'. Similarly, loss function used was 'Binary cross entropy' is preferred for binary classification. Adam optimizer achieved best result as it  handles noisy or sparse gradient problems much better than others. We found dropout rate '0.5' ideal for our problem case as increasing or decreasing it does not improve our results. Using learning rate, helps the model to adpat to the problem quickly. In our case we used '0.001' learning rate.
\begin{table}[width=.9\linewidth,cols=3,pos=h]
\caption{Hyper-parameters used for both datasets.}\label{tbl4}
\begin{tabular*}{\tblwidth}{@{} LLL@{} }
\toprule
\textbf{Parameters} & \textbf{UrduInAsmall} & \textbf{UrduInAlarge} \\
\midrule
Activation Function & sigmoid & sigmoid \\
Dropout & 0.5 & 0.5\\
Loss Function & binary crossentropy & binary crossentropy\\
Learning Rate & 0.001 & 0.001 \\
Optimizer & adam & adam \\
\bottomrule
\end{tabular*}
\end{table}
\subsection{Evaluation metrics}
The effectiveness of categorization models is frequently assessed by researchers using a variety of evaluation metrics. The Evaluation metrics used in our research are Precision, Recall, F1-score and accuracy.\\
A Precision metric counts how many correctly positive predictions were made. So the accuracy of class with minority can be measured using Precision. It is the ratio of :
\begin{equation}
Precision = \frac{TP}{TP + FP}
\end{equation}
where TP stands for True Positives i.e correct positive predictions and FP stands for false positives i-e incorrectly predicted positive. And (TP + FP) indicates total positive predictions.\\
Recall measures the proportion of positives that are correctly predicted among all possible positive predictions. Its ratio is:
\begin{equation}
Recall = \frac{TP}{TP + FN}
\end{equation}
where FN stands for false negatives it occurs when the model incorrectly predicts the negative class.
Precision and recall can be combined into one metric using F-Measure or F1-score, which covers both characteristics.Its can be measured as:
\begin{equation}
F1 - Score = 2 * \frac{Precision * Recall}{Precision + Recall}
\end{equation}
Accuracy in classification problems is used widely, due to the fact that it is a single metric that summarises the performance of model. It is the ratio of predictions that are predicted correctly by the model. It is as follows:
\begin{equation}
Accuracy = \frac{TP +TN}{TP +TN + FP +FN}
\end{equation}
where TN are those predictions that are correctly predicted negatives by the model.
\section{Results and Discussion}
This section explores the results from proposed model and other baseline models on two datasets and are discussed in detail. We addressed the identified research gaps and analyze the results with the evaluation metrics. The performance was observed in both datasets and impact of the Word2Vec word embedding is studied.\\
Our proposed model BiGRU with attention layer was first trained on both the datasets. It outperformed all other baseline DL models used in this study. The comparison of results is carried out by keeping in mind the size of dataset, evaluation measures, and the use of the embedding layer. We ran the experiments multiple times by altering the optimization parameters in order to achieve the best results from each model for perfect comparison.
\subsection{Comparison based on different factors}
\begin{table}[width=.9\linewidth,cols=6,pos=h]
\caption{Results Comparison of baseline Model with our proposed model on UrduInAsmall dataset without using Word2Vec layer.}\label{tbl5}
\begin{tabular*}{\tblwidth}{@{} LLLLLL@{} }
\toprule
\textbf{Model} & \textbf{Test Accuracy} & \textbf{Test Loss} & \textbf{F1-score} & \textbf{Precision} & \textbf{Recall}\\ 
\midrule
LSTM & 0.779 & 0.432 & 0.783 & 0.758 & 0.810 \\
Bi-LSTM & 0.773 & 0.545 & 0.763 & 0.787 & 0.740 \\
GRU & 0.770 & 0.493 & 0.768 & 0.763 & 0.773 \\
TCN & 0.770 & 0.442 & 0.743 & 0.825 & 0.676 \\
\textbf{BiGRU-A} & \textbf{0.789} & \textbf{0.496} & \textbf{0.781} & \textbf{0.797} & \textbf{0.766} \\
\bottomrule
\end{tabular*}
\end{table}
\begin{table}[width=.9\linewidth,cols=6,pos=h]
\caption{Results Comparison of baseline Model with our proposed model on UrduInAsmall dataset with using Word2Vec layer.}\label{tbl6}
\begin{tabular*}{\tblwidth}{@{} LLLLLL@{} }
\toprule
\textbf{Model} & \textbf{Test Accuracy} & \textbf{Test Loss} & \textbf{F1-score} & \textbf{Precision} & \textbf{Recall}\\ 
\midrule
LSTM & 0.726 & 0.483 & 0.726 & 0.715 & 0.783 \\
Bi-LSTM & 0.712 & 0.500 & 0.741 & 0.665 & 0.837 \\
GRU & 0.690 & 0.533 & 0.643 & 0.744 & 0.567 \\
TCN & 0.663 & 0.586 & 0.614 & 0.707 & 0.542 \\
\textbf{BiGRU-A} & \textbf{0.730} & \textbf{0.476} & \textbf{0.710} & \textbf{0.756} & \textbf{0.669} \\
\bottomrule
\end{tabular*}
\end{table}
\begin{table}[width=.9\linewidth,cols=6,pos=h]
\caption{Results Comparison of baseline Model with our proposed model on UrduInAlarge dataset without using Word2Vec layer.}\label{tbl7}
\begin{tabular*}{\tblwidth}{@{} LLLLLL@{} }
\toprule
\textbf{Model} & \textbf{Test Accuracy} & \textbf{Test Loss} & \textbf{F1-score} & \textbf{Precision} & \textbf{Recall}\\ 
\midrule
LSTM & 0.827 & 0.385 & 0.828 & 0.833 & 0.823 \\
Bi-LSTM & 0.825 & 0.431 & 0.808 & 0.847 & 0.773 \\
GRU & 0.810 & 0.493 & 0.799 & 0.807 & 0.791 \\
TCN & 0.807 & 0.459 & 0.806 & 0.772 & 0.844 \\
\textbf{BiGRU-A} & \textbf{0.842} & \textbf{0.367} & \textbf{0.827} & \textbf{0.866} & \textbf{0.792} \\
\bottomrule
\end{tabular*}
\end{table}
\begin{table}[width=.9\linewidth,cols=6,pos=h]
\caption{Results Comparison of baseline Model with our proposed model on UrduInAlarge dataset with using Word2Vec layer.}\label{tbl8}
\begin{tabular*}{\tblwidth}{@{} LLLLLL@{} }
\toprule
\textbf{Model} & \textbf{Test Accuracy} & \textbf{Test Loss} & \textbf{F1-score} & \textbf{Precision} & \textbf{Recall}\\ 
\midrule
LSTM & 0.748 & 0.511 & 0.752 & 0.708 & 0.803 \\
Bi-LSTM & 0.747 & 0.670 & 0.726 & 0.749 & 0.704 \\
GRU & 0.523 & 0.692 & 0.500 & 0.620 & 0.634 \\
TCN & 0.682 & 0.558 & 0.720 & 0.635 & 0.832 \\
\textbf{BiGRU-A} & \textbf{0.760} & \textbf{0.489} & \textbf{0.745} & \textbf{0.756} & \textbf{0.735} \\
\bottomrule
\end{tabular*}
\end{table}
Results obtained through our experiments are presented in Table 5 - 8. The `Model' column represents our baseline models and proposed model. Evaluation metrics are then presented in the next columns for respective DL models. Our suggested model BiGRU-A yeilds best performance i-e 84\% accuracy as compared with other models.We have compared these outcomes on the basis of following three factors:
\begin{itemize}
    \item \textbf{Dataset size}\\
    Table 5 and Table 6 represents the results obtained for small sized Urdu data set UrduInAsmall. Whereas, Table 7 and Table 8 represents the results obtained for large sized Urdu data set UrduInAlarge. If we compare the accuracy achieved by our suggested model for both dataset we can observe a visible increase in the value as the size of dataset increases. This shows that as we increase the training data, the DL model has more training examples to train. Hence, it can learn a lot better than the small data set training examples. Therefore, all the models yielded best accuracy performances for UrduInAlarge dataset when compared with UrduInAsmall dataset.
    \item \textbf{Effect of word embedding}\\
    Since word embedding can more clearly show the relationship and information between words, its application has been researched to enhance model performance \cite{jinbao2021text}. Table 5 - 8 shows the results from two datasets with or without using word embedding word2vec layer. Our results shows that using word embeddings has yielded poor performances whereas it achieves best results without using word embedding. This is because the Inappropriate class in our dataset contains a lot of swear words that are not included in pre-trained Word2Vec word embedding \cite{khan2021urdu}.
    \item \textbf{Model Comparison}\\
    Based on evaluation metrics shown in Table 5 - 8, visible difference in the performance of our model can be observed for two datasets. Precision, Recall and, F1 measure of our suggested model has outperformed all other models overall. Also, the loss of test data is calculated for all the models and our model has performed well in this metric.
\end{itemize}
\subsection{Discussion on Proposed Model}
The outcomes from all of the aggregations shown here demonstrate how the method described in this research offers a higher level of accuracy than the baseline models. Our hybrid Attention-based Bidirectional GRU (BiGRU-A) presents the best performance on both datasets with or without using word2ec layer. Figure 3 compares and visualizes the evaluation metrics of the baseline model and suggested architecture. Almost all the evaluation scores obtained by our proposed model are represented by a purple color bar and are better than the other DL algorithms used for comparison in this study. For balanced class datasets, evaluation metric accuracy and F1-score are good measures of efficiency. Precision and Recall are calculated in this study to present a comparison of outcomes with previous studies.\\
\begin{figure}
	\centering
		\includegraphics[scale=.50]{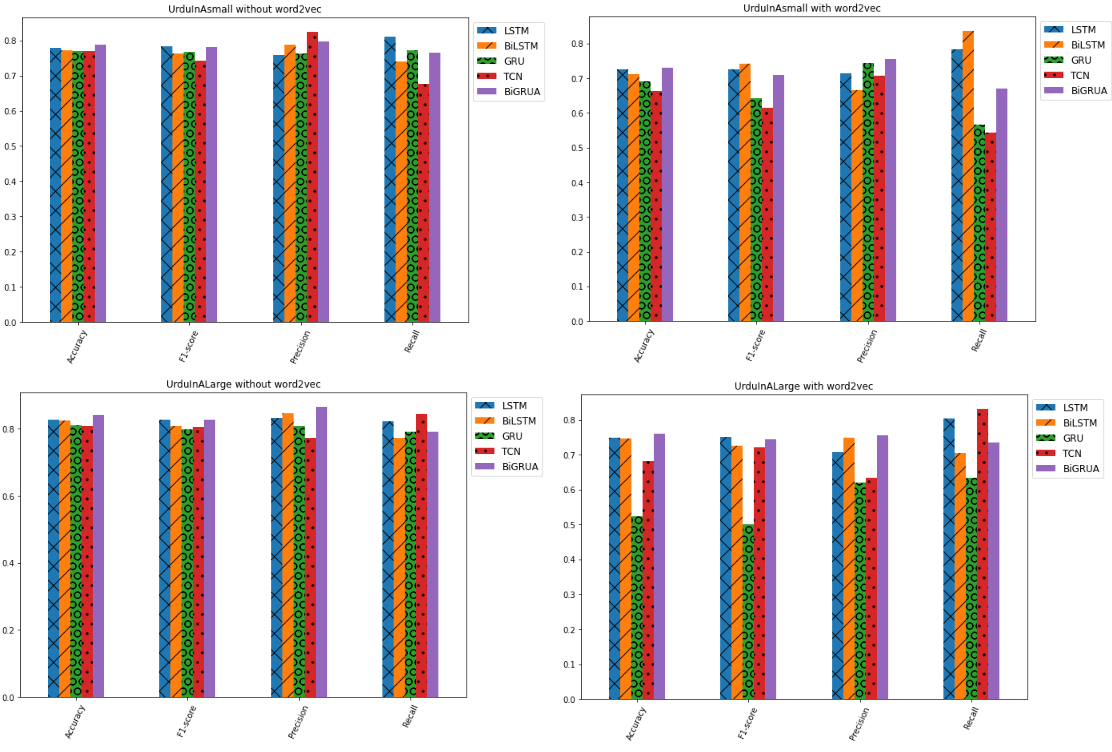}
	\caption{Evaluation Metrics Comparison of both data sets with or without Word2Vec}
	\label{FIG:3}
\end{figure}
Figure 4 presents a comparison of accuracy graphs of our proposed architecture on both datasets using word2Vec layer. The number of epochs set for this comparison are 50. Whereas, figure 5 gives a comparison of accuracy graphs of our proposed architecture on both datasets without using word2Vec layer. The number of epochs set for this comparison are 5. For UrduInAlarge the train and test accuracy is higher from the beginning representing the effect of increase in the dataset size. Whereas, for UrduInAsmall the initial accuracy is comparatively low and increases as more features are learned by the model.\\
Without using word2Vec layer, our model obtained 78.9\% accuracy on UrduInAsmall dataset and 84.2\% accuracy on UrduInlarge dataset which is compared with other models in table 5 and table 7. On the other hand, by including word2Vec layer, our model obtains 73\% accuracy on UrduInsmall dataset, and 76\% accuracy on UrduInAlarge dataset, which is compared with other models in table 6 and table 8.
\begin{figure}
	\centering
		\includegraphics[scale=.70]{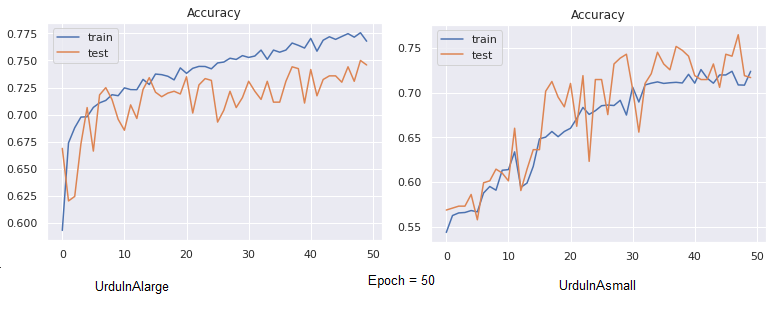}
	\caption{Accuracy Comparison of both data sets with Word2Vec}
	\label{FIG:4}
\end{figure}
\begin{figure}
	\centering
		\includegraphics[scale=.70]{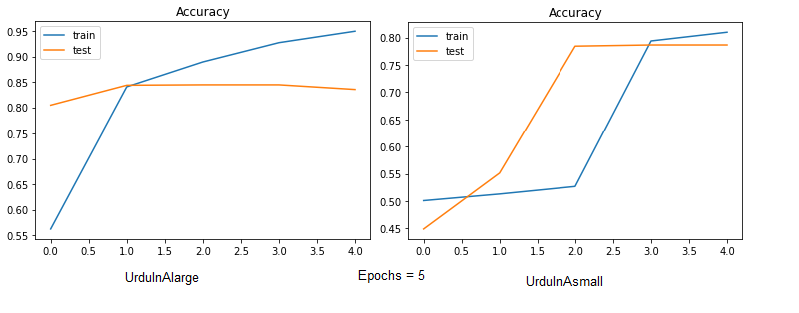}
	\caption{Accuracy Comparison of both data sets without Word2Vec}
	\label{FIG:5}
\end{figure}
To sum up, the technique utilized in the suggested hybrid model, BiGRU-A, is relatively simple. No feature selection technique was engaged throughout the experimentation process, giving it an edge over other DL models proposed in many studies. Our tests clearly shows that an attention layer can enable a model to grab specific important points in a sequence while computing its output. The attention layer also helps in handling long and variable-length sentences.\\
Hence our hybrid model not only reduces the cost and time of implementation but also enhance the performance of the DL model. This research work makes a contribution to the domain of inappropriate content identification in Urdu language using the DL techniques.
\subsection{Challenges of the Research}
The collection and annotation of the inappropriate content from Urdu text dataset was the first challenge faced in this research. When the Roman Urdu dataset was converted to Urdu script using online tool, the issue of incorrect translation of misspelled words was another challenge to face. People on social media not only have their  perspectives on spelling, but also use short forms as they wish. Human annotators can identify the real meaning behind short forms, however, there is a limit to what can be detected by automatic systems.\\
Another primary drawback is the potential for each model's training to proceed slowly. Each model must be trained once for each potential set of parameters. To achieve the best accuracy performance of all models different combinations of parameters, layers and neurons were utilized. Since it was difficult to test every potential combination, we move on to a few sets that ought to work well.
\section{Conclusion and Future work}
Many studies have explored the field of automatic inappropriate content detection in European or English language. But very few studies have considered investigating the inappropriate language detection in Urdu text using deep learning methods. We implemented bidirectional Gated Recurrent Unit along with attention layer on variable-sized datasets to detect inappropriate content in Urdu text language. A thorough comparative analysis been conducted with baseline deep learning models, and the impact of the word embedding layer is also studied deeply. Furthermore, are evaluated extensively using multiple metrics to make it a formal study in this field. The experimental results of our model revealed that it achieved overall high scores in every aspect, no matter the dataset size variations and the use of the word embedding layer. It yielded 84\% accuracy without using word embedding layer. Through our research, we have established that the use of the word embedding layer for inappropriate content detection decreases the efficiency of the model as this dataset contains many swear words that are not included in pre-trained word2Vec embedding. Moreover, the larger the size of the dataset more significant will be the performance of DL models.\\
This work can be extended by fine-tunning the word2Vec word embeddings to our dataset. We will also try different available word embeddings to further study the impact of embedding layer. Also, this study can be advanced further by implementing transfer learning approaches like BERT, Transformer, ML ensemble models to conduct additional analysis in the future.

\printcredits

\bibliographystyle{cas-model2-names}

\end{document}